\pdfoutput=1
\documentclass[11pt]{article}

\usepackage[]{naacl2021}

\usepackage{times}
\usepackage{latexsym}

\usepackage{subfiles}
\usepackage{url}
\usepackage{graphicx}
\usepackage{amsfonts}
\usepackage{amsmath}
\usepackage{amsthm}
\usepackage{subfig}
\usepackage{amssymb}
\usepackage{mathtools}
\usepackage{bbm}
\theoremstyle{definition}

\usepackage{algorithm}
\usepackage{algorithmic}
\usepackage{multirow}
\usepackage{caption}
\usepackage{booktabs}
\usepackage{xcolor}
\usepackage{ctable}

\usepackage[T1]{fontenc}
\usepackage[utf8]{inputenc}

\usepackage{microtype}

\title{AutoKG: Constructing Virtual Knowledge Graphs \\ from Unstructured Documents for Question Answering}

\author{Seunghak Yu, Tianxing He, and James Glass\\
  MIT Computer Science and Artificial Intelligence Laboratory, Cambridge, MA, USA \\
  \texttt{\{seunghak,cloudygoose,glass\}@csail.mit.edu} \\}

\begin{document}
\maketitle
\begin{abstract}

%We propose a method to automatically construct a knowledge graph that does not require to align the entities/relations in unstructured documents into human-annotated external knowledge sources and explore it to extract the desired information. 

%Knowledge graph is relevant to many NLP tasks, however, a major obstacle to its widespread adoption is the need to create a reliable domain-specific knowledge graph, which is time-consuming and expensive. A number of methods for constructing knowledge graphs without human intervention have been proposed, but the process of aligning to predefined information is essential.

Knowledge graphs (KGs) have the advantage of providing fine-grained detail for question-answering systems.  Unfortunately, building a reliable KG is time-consuming and expensive as it requires human intervention. To overcome this issue, we propose a novel framework to automatically construct a KG from unstructured documents that does not require external alignment. We first extract surface-form knowledge tuples from unstructured documents and encode them with contextual information. Entities with similar context semantics are then linked through internal alignment to form a graph structure. This allows us to extract the desired information from multiple documents by traversing the generated KG without a manual process. We examine its performance in retrieval based QA systems by reformulating the WikiMovies and MetaQA datasets into a tuple-level retrieval task. The experimental results show that our method outperforms traditional retrieval methods by a large margin.

%Knowledge graphs (KGs) are relevant to many NLP tasks, but building a reliable domain-specific KG is time-consuming and expensive. A number of methods for constructing KGs with minimized human intervention have been proposed, but still require a process to align into the human-annotated knowledge base. To overcome this issue, we propose a novel method to automatically construct a KG from unstructured documents that does not require external alignment and explore its use to retrieval QA tasks. To summarize our approach, we first extract knowledge tuples in their surface-form from unstructured documents, encode them using a pre-trained language model, and link the surface-entities via the encoding to form the graph structure.  We perform experiments with benchmark datasets such as WikiMovies and MetaQA. The experimental results show that our method can successfully create and search a KG with 18K documents and achieve 69.7\% hits@10 (close to an oracle model) on a query retrieval task.

%Unlike prior approaches, our method does not require prerequisite information, so it is able to represent relations for new (zero shot) classes of information.

\end{abstract}

%\htx{looking for a better term for "virtual" KG}

\section{Introduction}
\label{sec:intro}

Knowledge graphs (KGs) have been a key component of question-answering (QA) systems~\cite{hao2017end, chen2019bidirectional,deng2019multi,sun2019pullnet}. This is because a KG not only allows the QA system to utilize well-refined information through multi-hop inference, but also provides an interpretability mechanism to inform users about how an answer is extracted. However, using such well-refined structured knowledge has a scalability issue in acquiring information from various domains because of the high cost of constructing a knowledge graph, making it difficult to apply it to solving the problem for the latest emerging domains. To address this issue, many approaches have attempted to build KGs automatically~\cite{dong2014knowledge, bosselut2019comet, zhao2020complex}, but the steps required to align extracted knowledge with prior existing knowledge sources are necessary, so expert labor is still required. While there exists large general-purpose knowledge bases such as FreeBase~\cite{bollacker2008freebase} and Wikidata~\cite{vrandevcic2014wikidata}, these structured knowledge sources can easily become outdated as human knowledge is continually expanding, which makes it unsuitable for recent events.

An attractive way to reduce human labor is applying reading comprehension models on unstructured documents~\cite{chen2017reading, wang2018r, clark2018simple}.
It has the advantage of being able to understand unstructured documents without human labor, so it is inexpensive and can respond sensitively to the latest information. However, extracting the answer to a given question from a large pool of documents is a difficult task because there is a limit to the length of the context that can be understood by the model due to memory constraints. Most reading comprehension models are combined with a retrieval model because they focus on paragraph-level context and rely on the retrieval model to select candidates~\cite{chen2017reading, wang2019multi, feldmanmulti, karpukhin2020dense}. Therefore, the overall performance is bounded by the retrieved candidates. 

In this work, we aim to build a system combining advantages from both worlds.
We build a virtual KG from unstructured documents by extracting surface-form knowledge tuples from unstructured documents first, and then link them through internal alignment to form a graph structure. When given a question, we do multi-hop traversals on the virtual KG to extract desired information. To evaluate our retrieval based QA system, we reformulate two QA benchmark datasets into retrieval QA tasks. We find that our system significantly surpasses the performance of de-facto retrieval models which are used in most QA studies.

\section{Approach}
\label{sec:approach}

We propose a framework to automatically construct a virtual KG from unstructured documents, and during retrieval, traverse it to extract desired information from multiple documents. Given unstructured documents  $D=\{D_{i}\}$, where each document $D_{i}$ is a sequence of sentences $\{S_{j}\}$, we first create a KG  by extracting a set of knowledge tuples for each sentence $S{j}$ and link similar entities via contextual embeddings. Then, when given a query we traverse the generated graph multi-hops to find a tuple containing the answer. Note that it provides more fine-grained information than sentence-level retrievers in that each sentence can have multiple knowledge tuples. Our framework does not require external alignment to existing KB or any human intervention. Each of these two phases is described in detail in the following sections.

\subsection{Building a Virtual Knowledge Graph}
\label{sec:app_buildingkb}

Creating a graph from raw text is the key of this work. We use a three-step process to do this: \textit{Conversion}, \textit{Encoding}, and \textit{Surface-entity Linking}.

\textbf{Conversion:} We start by applying OpenIE ~\cite{saha2018open} to every sentence $S$ to generate a list of entity-relation triples. We will use the terms ``surface-entities'' ($se$) and ``surface-relation'' ($sr$), because they are not predefined entities/relations.
For each document $D$, we merge all triples extracted from each sentence into a list $T_D=\{(se^1_i, sr_i, se^2_i)\}$. Note that OpenIE extraction is imperfect: In many cases, it extracts something that is not a named entity like "it", "she", and extracted named entities contain noise.  In Table~\ref{tab:openie_example}, for example, ``\texttt{by Richard Donner}'' is extracted instead of the correct entity ``\texttt{Richard Donner}''. Moreoever, different surface-entities can refer to the same underlying entity. We address these problems in the next two steps.

\begin{table}
\small
\centering
\begin{tabular}{c|c|c}
\hline
\multicolumn{3}{l}{\textbf{Sentence:} The Goonies is an American film directed by} \\ 
\multicolumn{3}{l}{Richard Donner.} \\ \hline
\textbf{sf-entity1}                     & \textbf{sf-relation}               & \textbf{sf-entity2}                    \\ \hline
The Goonies                 & is                     & an American film           \\ \hline
an American film            & directed            & by Richard Donner             \\ \hline
\end{tabular}
\caption{\label{tab:openie_example}Example of the transformation from raw text to entity-relation triples by OpenIE. ``sf'' stands for ``surface''.}
\end{table}

\textbf{Encoding:} In the second step we utilize BERT \citep{devlinbert} to encode each surface-entity or surface-relation from extracted tuples $T_D$.  The challenge is to incorporate contextualized information into the encoding (e.g., the word ``Apple'' has different meanings in different contexts). Inspired by \cite{clarkbert}, we adopt the \textit{weighted embedding} technique, where the encoding of each surface-entity or relation is a simple weighted summation of the word embeddings and the embedding of \texttt{[CLS]} token when $T_D$ is fed into BERT. We denote the resulting encoding of $se$ as $se^{enc}$.

\textbf{Surface-entity Linking:} The third and final KG building step performs surface-entity linking, which creates a graph structure out of the extracted entity-relation triples in a document $D$. The goal is to link entities with same underlying concept together, for  example in Table~\ref{tab:openie_example}, a new surface-entity ``\texttt{the film}'' can be referring to ``\texttt{the Goonies}'' in some follow-up sentence, so the  relations with ``\texttt{the film}'' should also be applied to ``\texttt{the Goonies}''. We use an adaptive threshold on the cosine distance between the encodings of surface-entities to decide whether they should be linked.The intuition is that if there exists an $se_l$ that has high similarity to $se_i$, then the acceptable similarity threshold for $se_i$ should be higher. We denote the set of surface-entities linked to $se_i$ as $Link(se_i) = \{se_j : similarity(se^{enc}_i, se^{enc}_j) \geq \lambda * \max_{l\in E_D}(dist(se^{enc}_i, se^{enc}_l))\}$. Note that $E_D$ denotes the set of all surface-entities existing in document $D$. $\lambda$ is a hyper-parameter controlling the adaptive threshold, and we found that a setting of 0.6 works well in our experiments.
%_\Tilde{k}
%which is formulated below:
% \begin{equation}
% \begin{split}
%  Link(se_i) = \{se_j : cos(se^{enc}_i, se^{enc}_j) \geq \\ 
% \lambda * \max_{l\in E^D}(cos(se^{enc}_i, se^{enc}_l))\} %_\Tilde{k}
% \end{split}
% \end{equation}

To summarize, after the above three steps, each document has a list of extracted entity-relation triples $T_D=\{(se^1_i, sr_i, se^2_i)\}$, and each surface-entity or relation has a contextualized encoding. Within every document, each surface-entity $se_i$ is linked to $Link(se_i)$.

\subsection{Multi-hop KG Traversal and Retrieval}
\label{sec:traverse_vkg}

\begin{figure}
    \centering
    \includegraphics[width=0.45\textwidth]{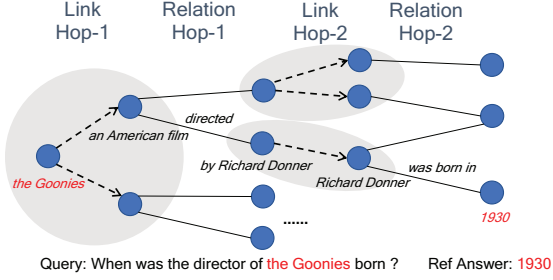}
    \caption{An illustration of a traversal on the constructed KG. Each node represents a surface entity, dashed arrows represent links between entities that exist in different tuples, and straight lines represent relations within tuples.}
    \label{fig:kg_traversal}
\end{figure}

We traverse the constructed KG to find relevant information to a given query $q$. The traversal algorithm is shown in Algorithm \ref{alg:graph_traverse} and we provide an illustration in Figure \ref{fig:kg_traversal}. To start the traversal, we select a set of \textit{seed} surface-entities in the virtual graphs as the start points. In most cases, we simply use the set of surface-entities that exist in the query $q$. If that set is empty, we encode $q$ with BERT as $q^{enc}$, and use the surface-entity whose encoding has the largest cosine-similarity with $q^{enc}$ as the seed entity. In this way, we can traverse in multiple documents that have seed entities. 

The output of the traversal will be a set of paths. We use the traversed triples to represent a \textit{path}: for example, $p=[se^1_1, sr_1, se^2_1, se^1_2, sr_2, se^2_2]$ is a 2-hop path. Starting with the seed surface-entities, we adopt an expand-and-prune strategy. For each hop, we first expand the current set of paths via the surface-entity linking, and then via the entity-relations triples. Since the number of active paths could grow exponentially during this expansion, we design an importance score to rate and prune the paths. For each path $p$, we first concatenate and feed it to 
BERT to get an encoding of this path $p^{enc}$. Then we use the cosine-similarity between $p^{enc}$ and $q^{enc}$ as the importance score. After the expansion of each hop, we only keep $B$ most relevant paths. In our experiments, we find that we only need to set $B$ to 10 to achieve good performance. Finally, we select the final top-$k$ paths as the output. 

\begin{algorithm}[h]
\small
   \caption{Traversing the KG}
   \label{alg:graph_traverse}
\begin{algorithmic}
   \STATE {\bfseries Input:} A set of seed surface-entities $\{se^{seed}_i\}$ and a query embedding $q_{enc}$
   \STATE {\bfseries Output:} A set of traversed paths %plural of path is paths! 
   
   \STATE For each seed surface-entity $se^{seed}_i$, initialize an empty path $p_i=[]$, and denote the set of paths as $P$. 
    
    \FOR {each hop}
    \STATE set $\hat{P}$ as an empty set
        \FOR {each $p \in P$}
            \STATE Let $se_p^{tail}$ be the last surface-entity in $p$
            \FOR {each $se_i \in Link(se_p^{tail})$}
                \FOR {each triple $(se_i, sr', se')$ in the KG which starts with $se_i$}
                    \STATE add $\hat{p} = p + [se_i, sr', se']$ to $\hat{P}$
                \ENDFOR
            \ENDFOR
        \ENDFOR 
    %\STATE // Pruning
    %        \STATE set $S_{p'}$ as a empty list
    %        \FOR {each $p' \in P'$}
    %            \STATE add $s_{p'} = cos(p'^{enc}, q_{enc})$ to $S_{p'}$
    %        \ENDFOR
    %\STATE prune $P'$ based on $S_{p'}$
    % htx: These seem reduntant to me
    \STATE Prune $\hat{P}$ based on $cos(\hat{p}^{enc}, q^{enc})$ with a beam size $B$
    \STATE set $\hat{P}$ as the new $P$
    \ENDFOR
    \STATE return $P$
\end{algorithmic}
\end{algorithm}

\label{sec:retrieval}

\section{Experimental Results}
\label{sec:experimental}

\subsection{Retrieval Performance Comparison}

\begin{table}[t]
\small
\begin{center}
\addtolength{\tabcolsep}{-1.0pt} 
\begin{tabular}{ c|c|cccc }
%\begin{tabular}{p{0.12\columnwidth}p{0.30\columnwidth}p{0.1\columnwidth}p{0.1\columnwidth}p{0.1\columnwidth}}
\hline
\textbf{Dataset} & \textbf{Model} & \textbf{H@1} & \textbf{H@3} & \textbf{H@5}\\
\hline
%\cite{miller2016key} & 68.30 &  & \\
%\hline
\multirow{5}{*}{WikiM} & BM25-sent & 12.94 & 39.55 & 46.78 \\
                       & BM25+RM3-tup & 15.03 & 20.98 & 25.78 \\ 
                       & QL+RM3-tup & 17.54 & 23.30 & 27.63 \\
                       & QL+RM3-sent & 25.71 & 40.96 & 47.83 \\
\cline{2-5}
            %& AutoKG (1-hop)& 43.78 & 59.40 &\bf{64.83} \\
            %& AutoKG (2-hop) & 46.24 & 53.84 & 57.54 \\
            %& AutoKG (3-hop) & \bf{54.54} & \bf{61.05} & 64.48 \\
            & AutoKG (3-hop) & \bf{54.54} & \bf{61.05} & \bf{64.48} \\
\hline
\multirow{5}{*}{Meta-2 }& BM25+RM3-tup & 10.77 & 13.43 & 16.13 \\ 
                        & QL+RM3-tup & 12.92 & 15.96 & 18.56 \\
                        & BM25+RM3-sent & 13.04 & 24.39 & 30.59 \\
                        & QL+RM3-sent & 18.93 & 30.33 & 36.15 \\
                        
\cline{2-5}
            % & AutoKG (1-hop)& 21.95 & 32.42 & 36.10 \\
            % & AutoKG (2-hop) & 30.73 & 40.40 &  45.29 \\
            % & AutoKG (3-hop) & \bf{36.77} & \bf{43.47} & \bf{47.04} \\
            & AutoKG (3-hop) & \bf{36.77} & \bf{43.47} & \bf{47.04} \\
\hline
\multirow{5}{*}{Meta-3} & BM25-sent & 6.28 & 18.54 & 24.16 \\
                        & BM25+RM3-tup & 7.58 & 10.05 & 11.91 \\
                        & QL+RM3-tup & 9.14 & 12.58 & 14.92 \\
                        & QL+RM3-sent & 11.52 & 20.34 & 25.64 \\
\cline{2-5}
            % & AutoKG (1-hop)& 18.17 & 27.39 & 31.62 \\
            % & AutoKG (2-hop) & 20.86 & 25.56 & 27.90 \\
            % & AutoKG (3-hop) & \bf{25.76} &\bf{32.86} & \bf{32.86} \\
            & AutoKG (3-hop) & \bf{25.76} & \bf{30.22} & \bf{32.86} \\
\hline
\end{tabular}
\end{center}
\caption{Comparison of retrieval performance.}
\label{Table:main_result}
\end{table}

\begin{table}[t]
\small
\begin{center}
\addtolength{\tabcolsep}{-1.0pt} 
\begin{tabular}{ c|c|cccc| }
%\begin{tabular}{p{0.12\columnwidth}p{0.2\columnwidth}p{0.1\columnwidth}p{0.1\columnwidth}p{0.1\columnwidth}}
\hline
\textbf{Dataset} & \textbf{Model} & \textbf{H@1} & \textbf{H@3} & \textbf{H@5}\\

\hline
\multirow{2}{*}{WikiM} & QL-doc & 55.46 & 70.24 & 75.68 \\
                       %& AutoKG & 54.54 & 61.05 & 64.48 \\
                       & AutoKG-doc & \bf{78.93} & \bf{85.10} & \bf{87.12}\\
\hline
\multirow{2}{*}{Meta-2 } & QL-doc & 36.76 & 53.24 & 60.11 \\
                        % & AutoKG  & 36.77 & 43.47 & 47.04 \\
                        & AutoKG-doc & \bf{57.82} & \bf{65.40} & \bf{67.57}\\
\hline
\multirow{2}{*}{Meta-3} & QL-doc & 21.07 & 34.79 & 44.44 \\
                        %& AutoKG & 25.76 & 30.22 & 32.86 \\
                        & AutoKG-doc & \bf{43.72} & \bf{47.33} & \bf{48.45}\\
\hline
\end{tabular}
\end{center}
\caption{Comparison with document-level systems.}
\label{Table:compare_document}
\end{table}

%\footnote{\url{https://research.fb.com/downloads/}}
%\footnote{\url{https://github.com/yuyuz/MetaQA}}

To quantitatively evaluate our model, we adopt two QA benchmark datasets: WikiMovies~\cite{miller2016key}, and MetaQA~\cite{zhang2018variational}. Since these datasets consist of pairs of questions and answers with related Wikipedia articles, we can leverage the entire set of articles in the dataset to build a KG and use our system to find the information that contains the correct answer to a given question. WikiMovies and MetaQA use the same 18,128 movie domain Wikipedia articles, but have different types of questions. We search the hyperparameter space using WikiMovies dev data (10K), and evaluate the model with the test data\footnote{Our framework does not involve training.}. The test data consists of 9,952  / 14,872 / 14,274 QA pairs respectively, for WikiMovies (1-hop), MetaQA (2-hop and 3-hop).

%1-hop (WikiMovies), 2-hop, and 3-hop (MetaQA) inference.

Finding a baseline to compare with our model is not straightforward because our system is different from conventional approaches in several ways: (1) Unlike existing automatic knowledge building methods that obtain results using external knowledge, we do not require external knowledge to align extractions.  (2) Unlike learning-based retrieval models~\cite{yilmaz2019applying, karpukhin2020dense} that require large amounts of labeled data to train the model, we do not require supervised training. Therefore, we focus on the retrieval aspect of our system, and compare with de-facto lexical retrieval methods to select candidate contexts, which are widely used in most open-domain QA studies~\cite{chen2017reading, wang2019multi, karpukhin2020dense}.

%\footnote{Please find Appendix~\ref{appsec:main_result} for all model results.}
We present the main results in Table~\ref{Table:main_result}, showing that our model outperforms de-facto retrieval methods by a large margin. AutoKG achieves 54.54 / 36.77 / 25.76 hits@1 for each WikiMovies (1-hop), MetaQA (2-hop and 3-hop) datasets, which is 2 to 3 times better than existing methods: BM25, QL, BM25+RM3, and QL+RM3. Hit@k~\cite{bordes2013translating} is the accuracy of top-k predicted paths containing the answer. BM25-tup is the result of applying BM25 at tuple-level after extracting triples through OpenIE for each document, and BM25-sent and -doc are retrieval results of each sentence- and document-level for original documents.

 %We also compare our system to the document-level system, as shown in Table~\ref{Table:compare_document}\footnote{Please find Appendix~\ref{appsec:doc_result} for all model results.}. 

%We also compare our system to document-level systems, as shown in Table~\ref{Table:compare_document}. Note that this is not a fair comparison because for the case of BM25-doc, it is considered a hit as long as the entire result documents contain an answer, while our system needs to locate the answer in the tuple. Additional experimental results and analysis can be found in Appendix.

Our system also surpasses the performance of document-level systems, as shown in Table~\ref{Table:compare_document}. Directly comparing our system to document-level systems is not appropriate, since document-level systems consider a hit when the resulting document contains the answer, whereas our system locates the answer in the tuple. For a fair comparison, we modify AutoKG to be a document-level system. When a tuple is retrieved, the system returns the document associated with that tuple. Additional results and analysis can be found in the Appendix.

\subsection{Quality Analysis of AutoKG}

Since the retrieval performance of the proposed method is bound by the quality of the resulting KG, we evaluate the coverage of AutoKG by comparing it to the human-annotated KB. Inspired by~\cite{gashteovski2018opiec}, we evaluate the coverage of AutoKG as follows: for each triple $({e_i}^1, r_i, {e_i}^2)$ in human-annotated KB, we consider the triple is covered by AutoKG, if any triple in AutoKG contains the tail entity ${e_i}^2$. We utilize human-annotated KB (based on OMDb~\cite{fritz2016omdb}) included in the WikiMovies dataset. This human-annotated KB contains knowledge that is not present in the documents of the dataset (e.g., IMDB ratings), so we measure all outcomes including or excluding such unreachable knowledge. 

Table~\ref{Table:coverage} shows the coverage of AutoKG compared to the human-annotated KB. ``Full-KB" is the result of measuring KB as it is, and ``Filtered-KB" is the result excluding them. We observe that the coverage of the proposed method is encouraging compared to the existing method obtained by running an engineered IE pipeline. This is because AutoKG can fully utilize the surface-form triples without external alignment.

%Note that our model utilizes entities and relationships in the surface-form, so it does not matter whether an exact match occurs. 

\begin{table}[t]
\small
\begin{center}
\begin{tabular}{c|c|c}
\hline
\textbf{Method} & \textbf{Full-KB (\%)} & \textbf{Filtered-KB (\%)} \\
\hline
\cite{miller2016key} &  69.35 & 83.69 \\
AutoKG & \textbf{75.16} & \textbf{90.00} \\
\hline
\end{tabular}
\end{center}
\caption{Coverage compared to human-annotated KB.}
\label{Table:coverage}
\end{table}

\section{Related Work}
\label{sec:related}

\paragraph{Automatic KG construction:} The most prevalent approach to create knowledge graphs from unstructured text involve developing a pipeline of NLP operations such as named entity recognition, entity linking and relationship extraction~\cite{gashteovski2018opiec, wu2019automatic}. These approaches require a predefined knowledge base to align the extracted entities or relationships~\cite{lin2016neural, zhou2016attention, zhang2019ernie, cao2020open}. Unlike conventional methods, our model can internally align surface-form triples extracted from unstructured documents, so no external knowledge base is required.

\paragraph{Graph based multi-hop retrievers:} In order to extract the desired information from multiple documents, \citet{sun2019pullnet} builds a question-relevant sub-graph from the knowledge base or text corpus to gather all the relevant information. This is similar to our approach in that it creates question-related sub-graphs, but differs from us in that it still requires a existing KB. \citet{das2019multi, asai2019learning} construct a Wikipedia graph using hyperlinks within the article to extract paragraphs related to the query. Therefore, their method contrasts with ours in that a human-annotated hyperlink is essential and the minimum unit of information to be searched is a paragraph. 

%\textbf{Graph based multi-hop retrievers:} In order to reason over documents and extract the desired information, it is necessary to extract information from multiple sentences or documents. To achieve this, \citet{sun2019pullnet} builds a question-relevant sub-graph from the knowledge base or text corpus to gather all the relevant information. This is similar to our approach in that it creates question-related sub-graphs, but differs from us in that it creates graphs using a predefined KB. \citet{das2019multi, asai2019learning} construct a Wikipedia graph using hyperlinks within the article to extract paragraphs related to the query. Therefore, their method contrasts with ours in that a human-annotated hyperlink is essential and the minimum unit of information to be searched is a paragraph. %\citet{dhingra2019differentiable}

\section{Conclusion}

We propose a novel framework to automatically build a knowledge graph from a pool of unstructured documents, without having to align resulting entities with external information. Our method outperforms existing de-facto retrieval models in retrieval-based QA tasks by a large margin. 

%As future work, we plan to improve multi-hop retrieval by introducing a trainable re-ranking module.

\section*{Acknowledgments}
Research was sponsored by the United States Air Force Research Laboratory and was accomplished under Cooperative Agreement Number FA8750-19-2-1000. The views and conclusions contained in this document are those of the authors and should not be interpreted as representing the official policies, either expressed or implied, of the United States Air Force or the U.S. Government. The U.S. Government is authorized to reproduce and distribute reprints for Government purposes notwithstanding any copyright notation herein.

\bibliography{anthology,naacl2021}
\bibliographystyle{acl_natbib}

\clearpage
\appendix
\section{Retrieval Performance Comparison}
\label{appsec:main_result}

Table~\ref{Table:main_result_app} shows the overall experimental results of retrieval performance comparison. We utilize the open-source information retrieval toolkit Anserini\footnote{\url{https://github.com/castorini/anserini}} for all our baseline experiments. All baseline models run with default parameters: BM25 ($k1=0.9, b=0.4$), Query Likelihood (Dirichlet smoothing parameter $\mu= 1000$), RM3 query expansion (the number of expansion terms $ nt = 10$, the number of expansion docs $nd = 10$, original query weight=$ w = 0.5$). 

The evaluation metric hit@k~\cite{bordes2013translating} is the accuracy of top-k predicted paths containing the answer. BM25-tup is the result of applying BM25 in tuple-level after extracting triples through OpenIE for each document, BM25-sent is the retrieval results at each sentence-level for the original documents, and BM25+RM3 shows the performance of BM25 with RM3 query expansion applied.

We observe that our model performs much better than all the de-facto retrieval models. In general, applying the RM3 query extension improves performance, and the query likelihood (QL) model performs better than the BM25.

\section{Comparison with document-level systems}
\label{appsec:doc_result}

Table~\ref{Table:document} shows the results of the overall experiment results comparing retrieval performance with document-level systems. All results are reported using the default parameters of each model as in previous experiments. Our model also outperforms document-level models. 

There exist interesting discoveries. First, unlike tuple- and sentence-level experiments, AutoKG (1-hop) performs better than AutoKG (3-hop). Second, RM3 query expansion is generally a factor that degrades performance. 

\section{Hyperparameter Search} 

We use the WikiMovies development dataset to find the optimal parameters of the proposed model based on hits@10 performance. There are two parameters we need to determine:

 \begin{itemize}
    \item $\lambda = \{x|x\in\mathbb{R}|0\leq x \le 1\}$ in surface-entity linking.
    \item  $ B = \{y|y \in \mathbb{Z}^+|y \geq 1\}$ in KG traversal. 
 \end{itemize}
 
 \begin{table}[t]
\begin{center}
\addtolength{\tabcolsep}{-3.0pt} 
\begin{tabular}{ c|c|cccc }
\hline
\textbf{Dataset} & \textbf{Model} & \textbf{H@1} & \textbf{H@3} & \textbf{H@5}\\
\hline
\multirow{11}{*}{WikiM} & BM25-tup & 5.08 & 17.97 & 24.69 \\
                       & QL-tup & 7.89 & 19.88 & 26.17 \\
                       & BM25+RM3-sent & 11.71 & 33.74 & 42.47 \\
                       & BM25-sent & 12.94 & 39.55 & 46.78 \\
                       & BM25+RM3-tup & 15.03 & 20.98 & 25.78 \\ 
                       & QL+RM3-tup & 17.54 & 23.30 & 27.63 \\
                       & QL-sent & 19.37 & 42.95 & 49.96 \\
                       & QL+RM3-sent & 25.71 & 40.96 & 47.83 \\
\cline{2-5}
            & AutoKG (1-hop)& 43.78 & 59.40 &\bf{64.83} \\
            & AutoKG (2-hop) & 46.24 & 53.84 & 57.54 \\
            & AutoKG (3-hop) & \bf{54.54} & \bf{61.05} & 64.48 \\
\hline
\multirow{11}{*}{Meta-2 } & BM25-tup & 9.72 & 14.09 & 16.88 \\
                         & BM25+RM3-tup & 10.77 & 13.43 & 16.13 \\ 
                         & QL-tup & 11.36 & 15.76 & 18.57 \\
                         & QL+RM3-tup & 12.92 & 15.96 & 18.56 \\
                         & BM25+RM3-sent & 13.04 & 24.39 & 30.59 \\
                         & QL+RM3-sent & 18.93 & 30.33 & 36.15 \\
                         & BM25-sent & 19.76 & 30.93 & 35.91 \\
                         & QL-sent & 22.07 & 33.51 & 38.45 \\
                        %  & BM25-tup & 7.43 & 11.43 & 14.23 \\
                        %  & BM25-sent & 9.78 & 17.29 & 20.93 \\
\cline{2-5}
             & AutoKG (1-hop)& 21.95 & 32.42 & 36.10 \\
             & AutoKG (2-hop) & 30.73 & 40.40 &  45.29 \\
             & AutoKG (3-hop) & \bf{36.77} & \bf{43.47} & \bf{47.04} \\
\hline
\multirow{11}{*}{Meta-3} & BM25-tup & 2.98 & 7.10 & 8.41 \\ 
                         & BM25+RM3-sent & 5.42 & 17.67 & 23.75 \\
                         & QL-tup & 5.72 & 11.74 & 15.12 \\
                         & BM25-sent & 6.28 & 18.54 & 24.16 \\
                         & BM25+RM3-tup & 7.58 & 10.05 & 11.91 \\ 
                         & QL+RM3-tup & 9.14 & 12.58 & 14.92 \\
                         & QL-sent & 10.61 & 21.37 & 27.12 \\
                         & QL+RM3-sent & 11.52 & 20.34 & 25.64 \\
\cline{2-5}
             & AutoKG (1-hop)& 18.17 & 27.39 & 31.62 \\
             & AutoKG (2-hop) & 20.86 & 25.56 & 27.90 \\
             & AutoKG (3-hop) & \bf{25.76} & \bf{30.22} & \bf{32.86} \\
\cline{2-5}

\hline
\end{tabular}
\end{center}
\caption{Comparison of retrieval performance.}
\label{Table:main_result_app}
\end{table}

\begin{table}[]
\begin{center}
\addtolength{\tabcolsep}{-3.0pt} 
\begin{tabular}{ c|c|cccc }
\hline
\textbf{Dataset} & \textbf{Model} & \textbf{H@1} & \textbf{H@3} & \textbf{H@5}\\
\hline
%\cite{miller2016key} & 68.30 &  & \\
%\hline
\multirow{5}{*}{WikiM}  & QL+RM3-doc & 43.36 & 60.75 & 69.93\\
                        & BM25+RM3-doc & 45.75 & 60.75 & 68.00\\
                        & BM25-doc & 54.80 & 69.24 & 74.50 \\
                        & QL-doc & 55.46 & 70.24 & 75.68 \\
\cline{2-5}
            & AutoKG (1-hop)& \bf{78.93} & \bf{85.10} & \bf{87.12}\\
            & AutoKG (2-hop)& 66.97 & 69.39 & 70.66 \\
            & AutoKG (3-hop) & 77.80 & 78.85 & 79.54 \\

\hline
\multirow{5}{*}{Meta-2 }& BM25+RM3-doc & 31.52 & 47.69 & 55.63\\
                        & QL+RM3-doc & 33.69 & 50.03 & 57.75 \\
                        & BM25-doc & 34.95 & 51.29 & 58.19 \\
                        & QL-doc & 36.76 & 53.24 & 60.11 \\
\cline{2-5}
             & AutoKG (1-hop)& \bf{57.82} & \bf{65.40} & \bf{67.57}\\
             & AutoKG (2-hop)& 57.37 & 60.36 & 61.97\\
             & AutoKG (3-hop)& 57.56 & 59.02 & 59.79 \\
\hline
\multirow{5}{*}{Meta-3} & QL+RM3-doc & 17.60 & 31.59 & 42.21 \\
                        & BM25+RM3-doc & 18.07 & 31.82 & 41.40\\
                        & BM25-doc & 20.56 & 34.10 & 43.02 \\
                        & QL-doc & 21.07 & 34.79 & 44.44 \\
\cline{2-5}
             & AutoKG (1-hop)& \bf{43.72} & \bf{47.33} & \bf{48.45}\\
             & AutoKG (2-hop)& 35.67 & 37.05 & 37.91\\
             & AutoKG (3-hop)& 43.49 & 44.14 & 44.57\\
\cline{2-5}
\hline
\end{tabular}
\end{center}
\caption{Comparison with document-level systems.}
\label{Table:document}
\end{table}

\begin{figure}[h]
  \centering\includegraphics[width=\columnwidth]{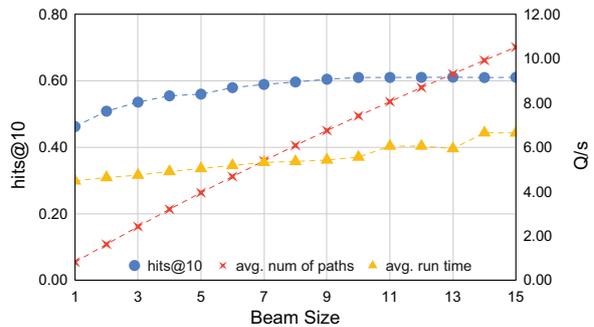}
  \caption{The effect of selecting beam size ($B$) on the model performance. $Q/s$ is a measure of how many queries have been processed in a second.}
  \label{Fig:beam_size}
\end{figure}

We select 0.6 for the linking threshold ($\lambda$) and 10 for the beam size ($B$) as the optimal parameters. Such small beam size $B$ effectively addresses the shortcoming of our method of exponentially expanding sub-graphs, making them practically applicable even when complex inference is required in multi-hop traversing. The effect of selecting $B$ on performance is shown in detail in Figure~\ref{Fig:beam_size}. We selected the optimal beam size as 10 considering performance and run time together.

\begin{table*}[]
\small
\begin{center}
\begin{tabular}{c|c|c|c|c}
\hline
\textbf{Methods} & \textbf{\# of topic} & \textbf{\# of tuple} & \textbf{tuple/topic} & \textbf{words/tuple}\\
\hline
OMDb & 17,342 & 101,800 & 5.87 & 11.64 \\
\cite{miller2016key} &  18,098 & 270,012 & 14.92 & 7.17 \\
AutoKG & \textbf{18,127} & \textbf{291,275} & \textbf{16.07} & \textbf{13.88} \\
\hline
\end{tabular}
\end{center}
\caption{Qualitative comparison of KGs.}
\label{Table:stat}
\end{table*}

\begin{table*}[]
\small
\centering
\begin{tabular}{c|l}
\hline
\textbf{Dataset} & \textbf{Example} \\
\hline
& Question : What does \textcolor{blue}{Jeremy Piven} act in?\\ 
& Golden Label : So Undercover, Keeping Up with the Steins, Just Write \\
WikiMovies & Predicted Paths (top-3) : \\
& 1. \textbf{White Palace film}: The movie features \textcolor{blue}{Jeremy Piven}.\\
(1-hop) & 2. \textbf{The Kingdom film}: starring Jason Bateman , with \textcolor{blue}{Jeremy Piven} is fictional. \\
& 3. \textbf{Very Bad Things}: (...) stars Christian Slater, with \textcolor{blue}{Jeremy Piven} in supporting roles.  \\
\hline
& Question : Who appeared in the same movie with \textcolor{blue}{Angie Everhart}?\\ 
& Golden Label : Erika Eleniak, Dennis Miller \\
MetaQA & Predicted Paths (top-3) : \\
& 1. Bordello of Blood: (...) starring \textcolor{blue}{Angie Everhart}. a 1996 comedy horror film starring \textit{Dennis Miller}.\\
(2-hop) & 2. Bordello of Blood: (...) starring \textcolor{blue}{Angie Everhart}. a 1996 comedy horror film starring \textit{Erika Eleniak}. \\
& 3. Bordello of Blood: (...) starring \textcolor{blue}{Angie Everhart}. a 1996 comedy horror film starring \textbf{Chris Sarandon}.
  \\
\hline
\end{tabular}
\caption{An example where the model found the correct answers but are considered incorrect due to the missing data labels. \textit{Italic} means the ground truth answers, \textbf{Bold} represents words that should have been correct.}
\label{Table:examples}
\end{table*}

\section{Quantitative Analysis of AutoKG}

Table~\ref{Table:stat} is the result of comparing basic statistics of KG generated by AutoKG with human-annotated KB and conventional pipeline information extraction (IE) method. Based on these statistics, we found some interesting facts.

First, considering the total number of documents (18,127) and the number of topics included in each KG, it can be seen that only the proposed method utilizes all documents in a given document pool to construct a KG. Human-annotated OMDb~\cite{fritz2016omdb} has the lowest number of 17,342, showing that manually generated KBs are easy to be outdated, which can cause issues in terms of coverage. In addition, in the existing IE pipeline method, we can see that information is dropped in the process of aligning the extracted tuple to the pre-existing KB.

Second, considering the number of tuples in KG and the average number of tuples for each topic, we can see that the proposed method extracts the most knowledge from unstructured documents. The IE-based model extracts more knowledge that is not in the human-annotated KB, but the proposed method does not filter knowledge with external alignment, so it can utilize more knowledge.

Finally, comparing the number of words used per tuple, we can see that the average word count in \citep{miller2016key} is less than human-annotate KB. This indicates that the system from \citep{miller2016key} could contain many tuples with invalid information due to the loss of information resulting from the filtering process, given that human-annotate KB likely has the most concise tuples.

\section{Qualitative Analysis of AutoKG} 

Example predictions are shown in Table~\ref{Table:examples}. We chose an example from two QA datasets, WikiMoives and MetaQA respectively. Each example shows the top-3 paths AutoKG predicts through the constructed KG for a given query. In the first example, we can see that AutoKG retrieves knowledge from three different documents and returns the correct answer. In the dataset, \textit{White Palace movie}, \textit{the Kingdom film} and \textit{Very Bad Things} are all separate documents. This shows that even though AutoKG constructs a graph of knowledge for each document, it is possible to extract information from multiple documents by simultaneously exploring multiple graphs using the seed entity present in the query as a starting point.

The second example shows that in a significant number of cases our model actually has the correct answer, but it is considered wrong because the reference label is incomplete. Therefore, we believe that the performance of our model is being underestimated.

\section{Implementation Details}

We remove the duplicate tuples created through OpenIE by using min-wise independent permutation locality sensitive hashing scheme~\cite{broder1997resemblance} with Jaccard similarity coefficient. If two sentences have 0.9 or more Jaccard similarity coefficient, it is considered the same and one of them is removed. By doing this, about 3\% of the tuples are removed from the extracted tuples. For the BERT model, we use huggingface's implementation~\cite{Wolf2019HuggingFacesTS}. Our method can be handled on a single GPU machine and has average run time as 5.55 query per GPU-second with NVIDIA V100 GPU.

\end{document}